\documentclass{svproc}
\usepackage{url}

\usepackage{graphicx}
\usepackage{amsmath,amssymb} 
\usepackage{color}
\usepackage{times}
\usepackage{epsfig}
\usepackage[utf8x]{inputenc}
\usepackage{mathtools}
\usepackage{makecell}
\usepackage{numprint}
\npthousandsep{\,}
\usepackage{algorithm}
\usepackage{multirow}
\usepackage{pifont}
\usepackage[width=122mm,left=12mm,paperwidth=146mm,height=193mm,top=12mm,paperheight=217mm]{geometry}
\newcommand{\xmark}{\ding{55}}%

\newcommand*{\suchthat}[1]{\left|\vphantom{#1}\right.}
\DeclareMathOperator*{\argmin}{arg\,min} 

\newcommand{\etal}{\textit{et al}.}

\newcommand{\eg}{\textit{e}.\textit{g}.}

\usepackage[breaklinks=true,bookmarks=false]{hyperref}

\begin{document}
\mainmatter              
\title{Efficient Nearest Neighbors Search for Large-Scale Landmark Recognition}

%
%
\author{Federico Magliani \and Tomaso Fontanini \and Andrea Prati}
%




%
\institute{IMP lab - University of Parma, 43124 Parma, Italy,\\
\email{federico.magliani@studenti.unipr.it},
\\ \texttt{http://implab.ce.unipr.it}}

\maketitle              

\begin{abstract}
The problem of landmark recognition has achieved excellent results in small-scale datasets. When dealing with large-scale retrieval, issues that were irrelevant with small amount of data, quickly become fundamental for an efficient retrieval phase. In particular, computational time needs to be kept as low as possible, whilst the retrieval accuracy has to be preserved as much as possible. In this paper we propose a novel multi-index hashing method called Bag of Indexes (BoI) for Approximate Nearest Neighbors (ANN) search. It allows to drastically reduce the query time and outperforms the accuracy results compared to the state-of-the-art methods for large-scale landmark recognition. 
It has been demonstrated that this family of algorithms can be applied on different embedding techniques like VLAD and R-MAC obtaining excellent results in very short times on different public datasets: Holidays+Flickr1M, Oxford105k and Paris106k.
\end{abstract}

\section{Introduction}

Landmark recognition is an emerging field of research in computer vision. In a nutshell, starting from an image dataset divided into classes, with each image represented by a feature vector, the objective is to correctly identify to which class a query image belongs.
This task presents several challenges: reaching high accuracy in the recognition phase, fast research time during the retrieval phase and reduced memory occupancy when working with a large amount of data.
The large-scale retrieval has recently become interesting because the results obtained in the majority of small-scale datasets are over the 90\% of the accuracy retrieval (\eg to Gordo \etal \cite{gordo2017end}). Searching the correct k nearest neighbors of each query is the crucial problem of large-scale retrieval because, due to the great dimension of data, a lot of distractors are present and should not be considered as possible query neighbors. 
In order to deal with large-scale datasets, an efficient search algorithm, that retrieves query results faster than n{\"a}ive brute force approach, while keeping a high accuracy, is crucial.
With an approximate search not all the returned neighbors are correct, then some are approximate, but they are typically still close to the exact neighbors. 
Usually, obtaining good results in the image retrieval task is strictly correlated with the high dimensionality of the global image descriptors, but on a large-scale version of the same problem is not advisable to use the same approach, due to the large amount of memory that would be needed. A possible solution is to first reduce the dimensionality of the descriptors, for example through PCA, and, then, apply techniques based on hashing functions for an efficient retrieval.

Following this strategy, this paper introduces a new multi-index hashing method called \textit{Bag of Indexes} (BoI) for large-scale landmark recognition based on Locality-Sensitive Hashing (LSH) and its variants, which allows to minimize the accuracy reduction with the growth of the data. The proposed method is tested on different public benchmarks using different embeddings in order to prove that is not an ad-hoc solution.

This paper is organized as follows. Section \ref{ref:rewo} introduces the general techniques used in the state of the art. Next, Section \ref{ref:boi} describes the proposed \textit{Bag of Indexes} (BoI) algorithm. Finally, Section \ref{ref:res} reports the experimental results on three public datasets: Holidays+Flickr1M, Oxford105k and Paris106k. Finally, concluding remarks are reported.

\section{Related work}\label{ref:rewo}

In the last years, the problem of landmark recognition was addressed in many different ways \cite{jegouVLAD} \cite{perronnin2010large} \cite{sivicBoW}. 
Recently, with the development of new powerful GPUs, the deep learning approach has shown its superior performance in many tasks of image retrieval \cite{tolias2015particular} \cite{gordo2016deep} \cite{babenko2014neural} \cite{yue2015exploiting}.  

Whenever the number of images in the dataset becomes too large, a Nearest Neighbor (NN) search approach to the landmark recognition task becomes infeasible, due to the well-known problem of the curse of dimensionality. Therefore, Approximate Nearest Neighbors (ANN) becomes useful, since it consists in returning a point that has a distance from the query equals to at most $c$ times the distance from the query to its nearest points, where $c > 1$.

One of the proposed techniques that allows to efficiently treat the ANN search problem is the Locality-Sensitive Hashing (LSH \cite{indyk1998approximate}), where the index of the descriptor is created through hash functions. LSH projects points that are close to each other into the same bucket with high probability. There are many different variants of LSH, such as E2LSH \cite{datar2004locality}, multi-probe LSH \cite{lv2007multi}, and many others. 

While LSH is a data-independent hashing method, there exist also data-dependent methods like Spectral Hashing \cite{weiss2009spectral}, which, however, is slower than LSH and therefore not appropriate for large-scale retrieval. In Permutation-Pivots index \cite{permutations}, data objects and queries are represented as appropriate permutations of a set of randomly selected reference objects, and their similarity is approximated by comparing their representation in terms of permutations. Product Quantization (PQ) \cite{jegou2011product} is used for searching local descriptors. It divides the feature space in disjoint subspaces and then quantizes each subspace separately. It pre-computes the distances and saves them in look-up tables for speeding up the search. Locally Optimized Product Quantization (LOPQ) \cite{kalantidis2014locally} is an optimization of PQ that tries to locally optimize an individual product quantizer per cell and uses it to encode residuals. Instead, FLANN \cite{muja2014scalable} is an open source library for ANN and one of the most popular for nearest neighbor matching. It includes different algorithms and has an automated configuration procedure for finding the best algorithm to search in a particular data set.

\section{Bag of Indexes}\label{ref:boi}

The proposed Bag of Indexes (BoI) borrows concepts from the well-known Bag of Words (BoW) approach. It is a form of multi-index hashing method \cite{greene1994multi} \cite{norouzi2012fast} for the resolution of ANN search problem. 

Firstly, following the LSH approach, $L$ hash tables composed by $2^\delta$ buckets, that will contain the indexes of the database descriptors, are created. 
The parameter $\delta$ represents the hash dimension in bits. The list of parameters of BoI and chosen values are reported in Table \ref{paramsSymbol} in Section \ref{adaptive_boi}.
Secondly, the descriptors are projected $L$ times using hashing functions. It is worth to note that this approach can be used in combination with different projection functions, not only hashing and LSH functions.
Finally, each index of the descriptors is saved in the corresponding bucket that is the one matching the projection result.

At query time, for each query, a BoI structure is created, that is a vector of $n$ weights (each corresponding to one image of the database) initiliazed to zero. Every element of the vector will be filled based on the weighing method explained in Section \ref{wm}. So, at the end of the projection phase, it is possible to make a coarse-grain evaluation of the similarity between the query image and the other images without calculating the Euclidean distance between them, but considering only their frequencies in the query buckets. Subsequently, at the end of the retrieval phase, the $\varepsilon$ elements of the vector with the highest weights are re-ranked according to their Euclidean distance from the query. The nearest neighbor is then searched only in this short re-ranked list. By computing the Euclidean distances only at the end of the retrieval phase and only on this short list (instead of computing them on each hash table like in standard LSH), the computational time is greatly reduced. Furthermore, this approach, unlike LSH, does not require to maintain a ranking list without duplicates for all the $L$ hash tables. 
The detailed analysis of the memory occupation of BoI is reported in Section \ref{ref:res}.

\begin{figure}[hbt]
\centering
\includegraphics[width= 9cm]{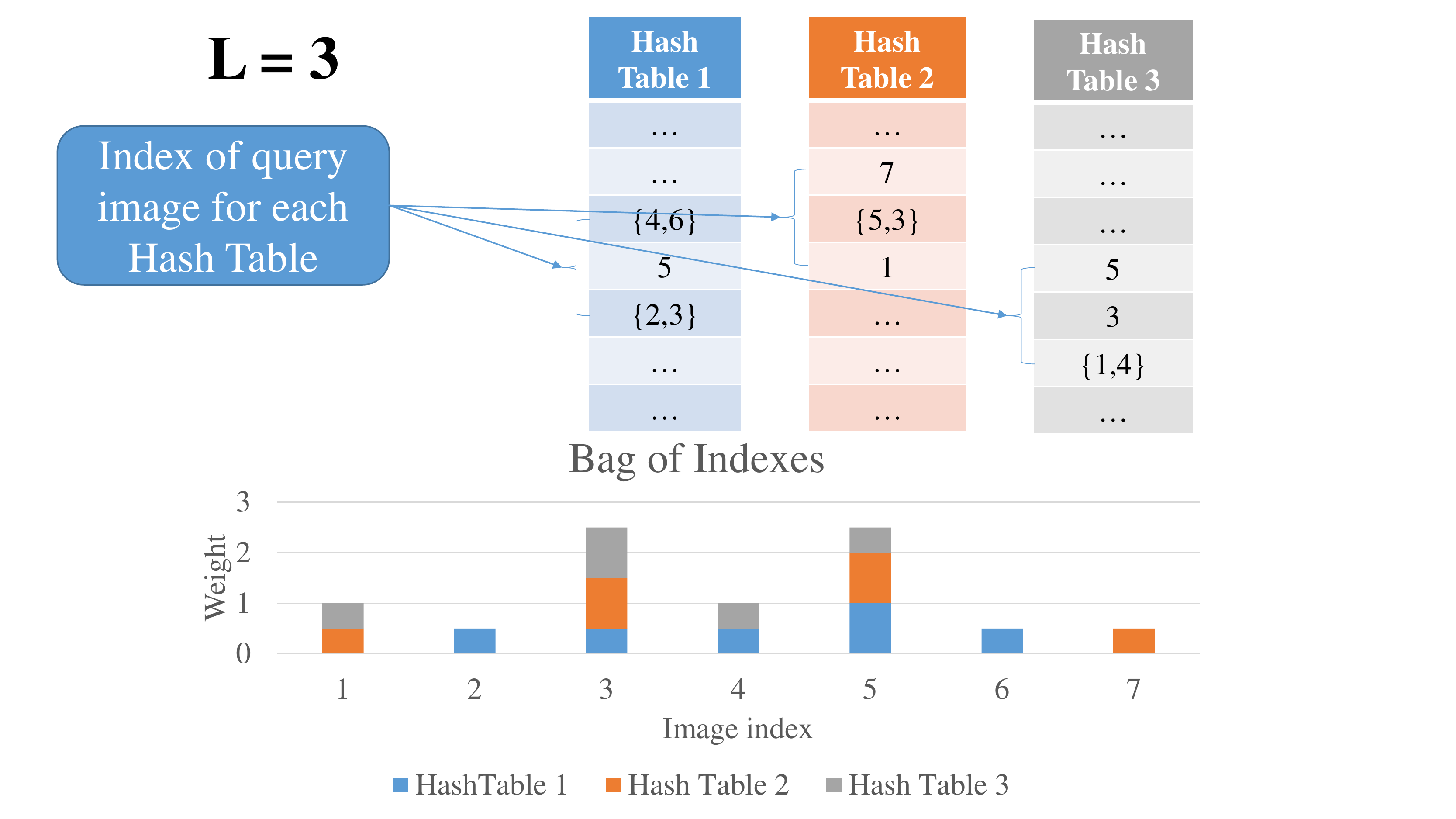}
\caption{Overview figure of the retrieval through BoI multi-probe LSH.}
\label{boiScheme}
\end{figure}

\subsection{Weighing metric}\label{wm}

As previously reported, BoI can be used in combination with different hashing functions.
When used with baseline LSH, the corresponding bucket of the query image will be checked. In this case, even thought it is faster than LSH, the accuracy suffers a significant loss.
Conversely, when BoI is combined with multi-probe LSH, also the $l$-neighboring buckets are considered.

The $l$-neighbors are the buckets that have a Hamming distance less than or equal to $l$ from the hashed value of the query, which corresponds to the query bucket. The weights for the any value of $l$ are chosen as follows:

\begin{equation}
 w(i,q,l) =
  \begin{cases}
    \frac{1}{2^{H(i,q)}}  & \quad \text{if } {H(i,q)} \leq l\\
   0  & \quad \text{otherwise}
  \end{cases}
  \label{weightsEq}
\end{equation}
\noindent where $i$ is a generic bucket, $q$ is the query bucket and $H(i,q)$ is the Hamming distance between $i$ and $q$.
The BoI multi-probe LSH approach increases the number of buckets considered during the retrieval and, thus, the probability of retrieving the correct result, by exploiting the main principle of LSH that similar objects should fall in the same bucket or in the ones that are close to it. However, even if we want to account for some uncertainty in the selection of the correct bucket, we also want to weight less as soon as we move farther from the ``central" bucket. 

Fig. \ref{boiScheme} shows an exemplar overview of the BoI computation. With $L = 3$ hash tables and 1-neighbours (i.e., $l=1$), a query can be projected in different buckets. The corresponding weights (see eq. \ref{weightsEq}) are accumulated in the BoI (see the graph on the bottom of the image). Only the $\epsilon$ images with the highest weights are considered for the last step (re-ranking) for improving the recall.

\subsection{BoI adaptive multi-probe LSH}\label{adaptive_boi}

This BoI multi-probe LSH approach has the drawback of increasing the computational time since it also needs to search in neighboring buckets (which are $\sum_{i=0}^{l} {{\log_2\delta}\choose{i}}$, being $\delta$ the hash dimension). To mitigate this drawback, we introduce a further variant, called \textit{BoI adaptive multi-probe LSH}. 
The main idea of this approach is to iteratively refine the search bucket space, by starting with a large number of neighboring buckets $\gamma_0$ (e.g., 10) and slowly reduce $\gamma$ when the number of hash tables increases. This adaptive increase of focus can, on the one hand, reduce the computational time and, on the other hand, reduce the noise. In fact, at each iteration, the retrieval results are supposed to be more likely correct and the last iterations are meant to just confirm them, so there is no need to search on a large number of buckets. In order to avoid checking the same neighbors during different experiments, the list of neighbors to check is shuffled randomly at each experiment.

Two different techniques for the reduction of the number of hash tables are evaluated:
\begin{itemize}
\item  \textit{linear}: the number of neighboring buckets $\gamma$ is reduced by 2 every 40 hash tables, i.e.:
\begin{equation}
\resizebox{.5 \textwidth}{!} 
{$
\gamma_i  =
  \begin{cases}
   \gamma_{i-1} - 2  & \quad \text{if } {i=\{\Delta_1,\dots,k_1\Delta_1\}}\\
    \gamma_{i - 1}  & \quad \text{otherwise}
   \end{cases}
$}
\label{eq:linear}
\end{equation}
with $i=\{1, \dots, L\}, \quad \Delta_1 = 40 \quad$ and $\quad k_1 : k_1\Delta_1 \le L$
\item \textit{sublinear}: the number of neighboring buckets $\gamma$ is reduced by 2 every 25 hash tables, but only after the first half of hash tables, i.e.:
\begin{equation}
\resizebox{.7 \textwidth}{!} 
{$
\gamma_i  =
  \begin{cases}
    \gamma_{i - 1}  & \quad \text{if } {i} \leq L/2\\
   \gamma_{i-1} - 2  & \quad \text{if } {i=\{L/2, L/2+\Delta_2,\dots,L/2+k_2\Delta_2\}}\\
    \gamma_{i - 1}  & \quad \text{otherwise}  
   \end{cases}
$}
\label{eq:sublinear}
\end{equation}
with $i=\{1, \dots, L\}, \quad \Delta_2= 25 \quad$ and $\quad k_2 : L/2+k_2\Delta_2 \le L$
\end{itemize}

\begin{table}[tbh]
 \centering
    \begin{tabular}{|c|c|c|}
    \hline
    \textbf{Symbol} & \textbf{Definition} & \textbf{Chosen value} \\ \hline
    $n$ & number of images & - \\ \hline
     $\delta$ & hash dimension & $2^8 = 256$ \\ \hline
    L & number of hash tables & 100\\ \hline
    $\gamma_0$ &  initial gap & 10 \\ \hline
   $l$ & neighbors bucket & 1-neighbors \\ \hline  
   $\varepsilon$ & elements in the re-ranking list & 250 \\ \hline
   - & reduction & sublinear \\ \hline
	\end{tabular}
\caption{Summary of notation.}
\label{paramsSymbol}
\end{table}

The proposed approach contains several parameters. Their values were chosen after an extensive parameter analysis (out of the scope of this paper) and summary of notation is reported in Table \ref{paramsSymbol}.
$L$, $\delta$ and $l$ should be as low as possible since they directly affect the number of buckets $\mathcal{N}_q^l$ to be checked and therefore the computational time at each query $q$, as follows:
\begin{equation}
\mathcal{N}_q^l=L\sum_{i=0}^{l} {{\gamma_i}\choose{i}}=
L\sum_{i=0}^{l}\frac{\left(\gamma_i\right)!}{i!\left(\gamma_i-i\right)!}
\end{equation}
\noindent where $\gamma_i=\gamma_0=\log_2\delta, \forall i$ for standard BoI multi-probe LSH, whereas, in the case of BoI adaptive multi-probe LSH, $\gamma_i$ can be computed using the eqs. \ref{eq:linear} or \ref{eq:sublinear}.

\section{Experimental results}\label{ref:res}

The proposed approach has been extensively tested on public datasets in order to evaluate the accuracy against the state of the art. 

\subsection{Datasets and evaluation metrics}

The performance is measured on three public image datasets: Holidays+Flickr1M, Oxford105k and Paris106k as shown in Table \ref{ds}. 

\begin{table*}
\centering
\setlength{\tabcolsep}{6pt}
    \begin{tabular}{|c|c|c|}
    \hline
     \textbf{Dataset} & \textbf{Size} &  \textbf{Query images} \\ \hline
     Holidays \cite{Holidays} + Flickr1M & \numprint{1001491} & 500 \\ \hline
     Oxford105k \cite{Oxford} & \numprint{105063} & 55 \\ \hline
     Paris106k \cite{Paris} & \numprint{106392} & 55 \\ \hline
    \end{tabular}
    \caption{Datasets used in the experiments}
        \label{ds}
    \end{table*}

\textbf{Holidays} \cite{Holidays} is composed by 1491 images representing the holidays photos of different locations, subdivided in 500 classes. The database images are 991 and the query images are 500, one for every class.

\textbf{Oxford5k} \cite{Oxford} is composed by 5062 images of Oxford landmarks. The classes are 11 and the queries are 55 (5 for each class).

\textbf{Paris} \cite{Paris} is composed by 6412 images of landmarks of Paris, France. The classes are 11 and the queries are 55 (5 for each class).

\textbf{Flickr1M} \cite{huiskes2008mir} contains 1 million Flickr images used as distractors for Holidays, Oxford5k and Paris6k generating Holidays +Flickr1M, Oxford105k and Paris106k datasets.


\textbf{Evaluation}. Mean Average Precision (mAP) was used as metrics for accuracy.

\textbf{Distance}.  $L_2$ distance was employed to compare query images with the database.

\textbf{Implementation}. All experiments have been run on 4 separate threads. The CNN features used for the creation of locVLAD \cite{maglianiLocVLAD} descriptors are calculated on a NVIDIA GeForce GTX 1070 GPU mounted on a computer with 8-core and 3.40GHz CPU. 
\subsection{Results on Holidays+Flickr1M datasets}

This section reports the results of our approach, by adding to the Holidays dataset a different number of distractors, obtained from the Flickr1M dataset. All the experiments have been conducted several times and a mean has been computed in order to eliminate the randomness of the Gaussian distribution used in the hashing function. The embeddings used are locVLAD descriptors \cite{maglianiLocVLAD}, while the features are extracted from the layer mixed8 of Inception V3 network \cite{szegedy2016rethinking} that is a CNN pre-trained on the ImageNet \cite{deng2009imagenet} dataset. The vocabulary used for the creation of locVLAD descriptors is calculated on Paris6k.


\begin{table}[htb]
\centering
\setlength{\tabcolsep}{6pt}
    \begin{tabular}{| c | c | c | c | }
    \hline
    \multirow{2}{*}{\textbf{Method}} & \multirow{2}{*}{\textbf{$\epsilon$}} &   \multicolumn{2}{c|}{\textbf{Holidays+Flickr1M}} \\ 
    \cline{3-4} &  & \textbf{mAP} & \textbf{avg retrieval time} \\ \hline
     LSH* & 250 & 86.03\% & \numprint{3103} \\ \hline
    Multi-probe LSH*  (L = 50) & 250 & 86.10\% & \numprint{16706} \\ \hline
    PP-index*  \cite{permutations} & 250 & 82.70\% & \numprint{2844} \\ \hline
    LOPQ \cite{kalantidis2014locally} & 250  & 36.37\% & 4 \\ \hline
    FLANN  \cite{muja2014scalable} & 250 & 83.97\% & 995 \\ \hline
    BoI LSH & 250 & 78.10\% & 5 \\ \hline
    BoI multi-probe LSH & 250 & 85.16\% & 12 \\ \hline
    BoI adaptive multi-probe LSH & 250 & 85.35\% & 8\\ \hline
 \hline
    PP-index* \cite{permutations} & 10k  & 85.51\% & \numprint{15640} \\ \hline
    LOPQ \cite{kalantidis2014locally} & 10k  & 67.22\% & 72 \\ \hline
    FLANN \cite{muja2014scalable} & 10k  & 85.66\% & \numprint{1004} \\ \hline
    BoI adaptive multi-probe LSH & 10k & \textbf{86.09\%} & \textbf{16} \\ \hline
   \end{tabular}
        \caption{Results in terms of mAP and average retrieval time in msec on Holidays+Flickr1M. * indicates our re-implementation.}

        \label{results1M}
\end{table}


Table \ref{results1M} summarizes the results on Holidays+Flickr1M dataset in terms of mAP and average retrieval time (msec). 
The first experiments evaluated only the top $\epsilon = 250$ nearest neighbors. 



\begin{figure}[hbt]
\centering
\includegraphics[width= 9cm]{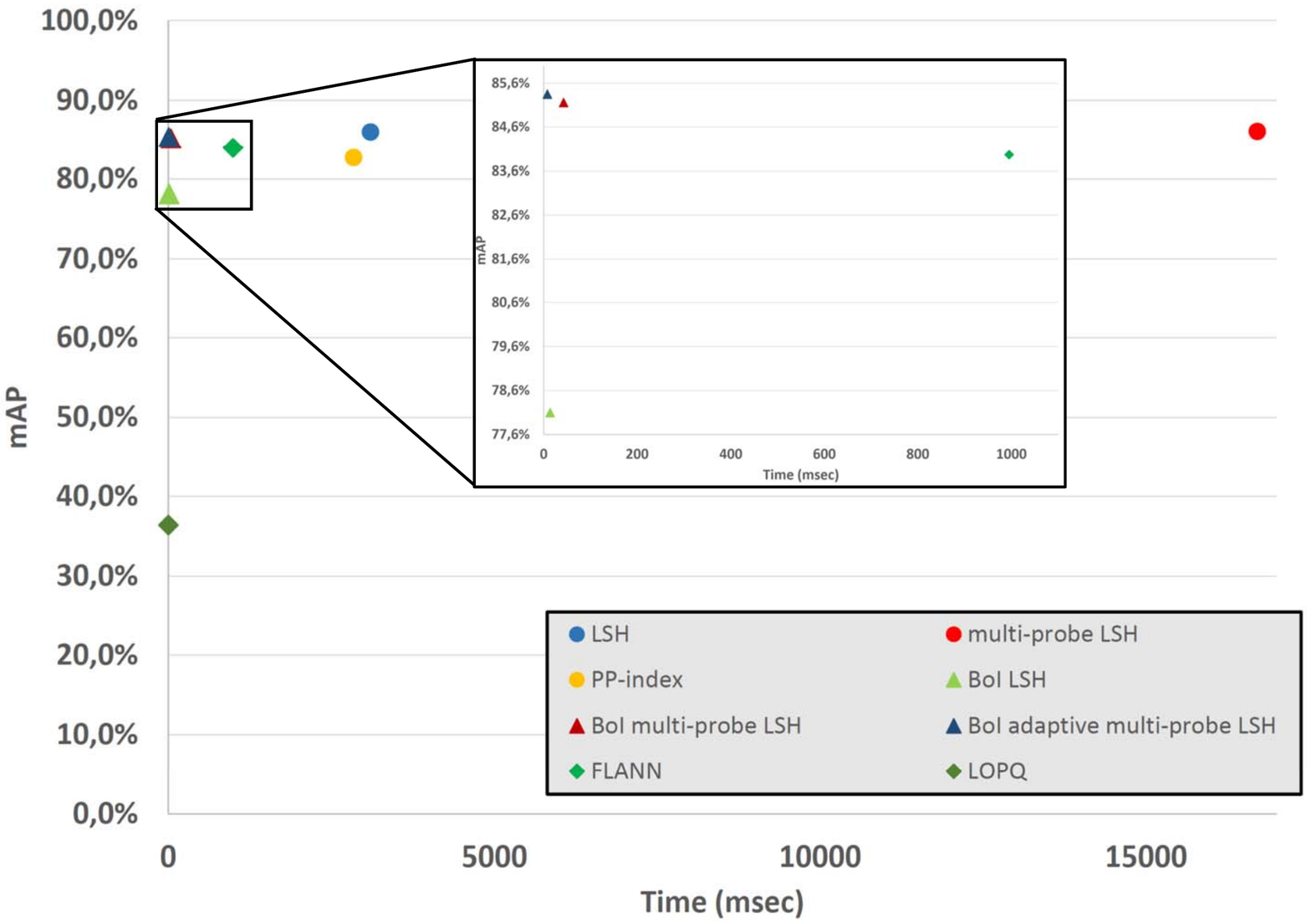}
\caption{Relationship between time and accuracy on Holidays+Flickr1M with different approaches.}
\label{timeacc}
\end{figure}

LSH and multi-probe LSH achieve excellent results, but with an huge retrieval time. Also PP-index \cite{permutations} needs more than 3 seconds for a query to retrieve the results. 
LOPQ\cite{kalantidis2014locally} and FLANN \cite{muja2014scalable} reach poor results on large-scale retrieval.  LOPQ reached 36.37\%, while FLANN achieved 83.97\%. However, while query time for LOPQ is pretty low, FLANN is not able to keep the query time low. It is worth saying that both LOPQ and FLANN has been tested using the available codes from authors and reported results correspond to the best found configuration of parameters. Given the significantly low (especially for LOPQ) performance in accuracy, further experiments have been conducted for LOPQ, FLANN, as well as PP-index and our method by increasing $\epsilon$ from 250 to 10k. As foreseeable, all the accuracy results improved with respect to $\epsilon = 250$ (LOPQ increases from 36.37\% to 67.22\%), but the proposed BoI adaptive multi-probe LSH method still outperforms all the others. Moreover, our method still results to be faster than the others (LOPQ is fast like ours, but with lower accuracy, while PP-index and FLANN are slightly lower in accuracy, but much slower).


Overall speaking, our proposal outperforms all the compared methods in the trade-off between accuracy and efficiency. To better highlight this, Fig. \ref{timeacc} shows jointly the mAP (on y-axis) and the average query time (on x-axis). The best trade-off has to be found in the upper left corner of this graph, i.e. corresponding to high accuracy and low query time. All the BoI-based methods clearly outperform the other methods.

Regarding the memory footprint of the algorithm for 1M images with 1M descriptors of 128D (float = 4 bytes), brute-force approach requires 0.5Gb (1M x 128 x 4). LSH needs only 100 Mb: 1M indexes for each of the L=100 hash tables, because each indexes is represented by a byte (8 bit) and so 1M indexes x 100 hash tables x 1 byte = 100Mb. The proposed BoI only requires additional 4 Mb to store 1M weights. 

\subsection{Results on Oxford105k and Paris106k datasets}

Since our goal is to execute large-scale retrieval for landmark recognition, we have also used the Oxford105k and Paris106k datasets.
In this case, all the methods are tested using R-MAC descriptors, fine-tuned by Gordo \textit{et al.} \cite{gordo2017end}, since VLAD descriptors are demonstrated to be not suited for these datasets \cite{liu2007survey}.
\begin{table}[htb]

\centering
\setlength{\tabcolsep}{6pt}
    \begin{tabular}{| c | c | c | c | c | c |}
    \hline
    \multirow{2}{*}{\textbf{Method}} & \multirow{2}{*}{\textbf{$\epsilon$}} &  \multicolumn{2}{c|}{\textbf{Oxford105k}} &  \multicolumn{2}{c|}{\textbf{Paris106k}} \\ 
    \cline{3-6} & & \textbf{mAP} & \textbf{avg ret. time} & \textbf{mAP} & \textbf{avg ret. time} \\ \hline
    LSH* & 2500 & 80.83\% & 610 & 86.50\% & 607  \\ \hline
    PP-index* \cite{permutations} & 2500 & 81.89\% & 240 & 88.14\% & 140 \\ \hline
    LOPQ \cite{kalantidis2014locally} & 2500 & 71.90\% & 346 & 87.47\% & 295 \\ \hline
    FLANN \cite{muja2014scalable} & 2500 & 70.33\% & 2118 & 68.93\% & 2132  \\ \hline
    BoI adaptive multi-probe LSH & 2500 & 81.44\% & 12 & 87.90\% & 13 \\ \hline
    \hline
    PP-index* \cite{permutations} & 10k & 82.82\% & 250 & 89.04\% & 164 \\ \hline
    LOPQ \cite{kalantidis2014locally} & 10k & 69.94\% & 1153 & 88.00\% & 841 \\ \hline
    FLANN \cite{muja2014scalable} & 10k & 69.37\% & 2135 & 70.73\% & 2156  \\ \hline
    BoI adaptive multi-probe LSH & 10k & \textbf{84.38\%} & \textbf{25} & \textbf{92.31\%} & \textbf{27}  \\ \hline

    \end{tabular}
        \caption{Results in terms of mAP and average retrieval time (msec) on Oxford105k and Paris106k. * indicates our re-implementation of the method.}
        \label{results100k}
\end{table}

Table \ref{results100k} show the mAP and the average retrieval time. Using $\epsilon = 2500$, the proposed approach obtained slightly worse results than PP-index, but resulted one order of magnitude faster in both datasets. When more top-ranked images are used ($\epsilon = 10k$), BoI adapative multi-probe LSH obtained the best results and with lower query time.
Furthermore, LOPQ \cite{kalantidis2014locally} works better on Paris106k than Oxford105k, while FLANN \cite{muja2014scalable} performs poorly on both datasets. 
\section{Conclusions}

In this paper, a novel multi-index hashing methods called Bag of Indexes (BoI) for approximate nearest neighbor search problem is proposed. This method demonstrated an overall better trade-off between accuracy and speed w.r.t. state-of-the-art methods on several large-scale landmark recognition datasets. Also, it works well with different embedding types (VLAD and R-MAC). 
The main future directions of our work will be related to reduce the dimension of the descriptor in order to speed the creation of bucket structure  and to adapt the proposed method for dataset with billions of elements. 

\textbf{Acknowledgments}. This work is partially funded by Regione Emilia Romagna under the “Piano triennale alte competenze per la ricerca, il trasferimento tecnologico e l’imprenditorialità”.


\bibliographystyle{splncs_srt}
\bibliography{egbib}

\begin{thebibliography}{10}

\bibitem{babenko2014neural}
Babenko, A., Slesarev, A., Chigorin, A., Lempitsky, V.:
\newblock Neural codes for image retrieval.
\newblock In: European conference on computer vision, Springer (2014)  584--599

\bibitem{permutations}
Chavez, E., Figueroa, K., Navarro, G.:
\newblock Effective proximity retrieval by ordering permutations.
\newblock IEEE transactions on Pattern Analysis and Machine Intelligence
  \textbf{30} (2008)  1647--1658

\bibitem{datar2004locality}
Datar, M., Immorlica, N., Indyk, P., Mirrokni, V.S.:
\newblock Locality-sensitive hashing scheme based on p-stable distributions.
\newblock In: Proceedings of the twentieth annual symposium on Computational
  geometry, ACM (2004)  253--262

\bibitem{deng2009imagenet}
Deng, J., Dong, W., Socher, R., Li, L.J., Li, K., Fei-Fei, L.:
\newblock Imagenet: A large-scale hierarchical image database.
\newblock In: Computer Vision and Pattern Recognition, 2009. CVPR 2009. IEEE
  Conference on, IEEE (2009)  248--255

\bibitem{gordo2016deep}
Gordo, A., Almaz{\'a}n, J., Revaud, J., Larlus, D.:
\newblock Deep image retrieval: Learning global representations for image
  search.
\newblock In: European Conference on Computer Vision, Springer (2016)  241--257

\bibitem{gordo2017end}
Gordo, A., Almazan, J., Revaud, J., Larlus, D.:
\newblock End-to-end learning of deep visual representations for image
  retrieval.
\newblock International Journal of Computer Vision \textbf{124}(2) (2017)
  237--254

\bibitem{greene1994multi}
Greene, D., Parnas, M., Yao, F.:
\newblock Multi-index hashing for information retrieval.
\newblock In: Foundations of Computer Science, 1994 Proceedings., 35th Annual
  Symposium on, IEEE (1994)  722--731

\bibitem{huiskes2008mir}
Huiskes, M.J., Lew, M.S.:
\newblock The mir flickr retrieval evaluation.
\newblock In: Proceedings of the 1st ACM international conference on Multimedia
  information retrieval, ACM (2008)  39--43

\bibitem{indyk1998approximate}
Indyk, P., Motwani, R.:
\newblock Approximate nearest neighbors: towards removing the curse of
  dimensionality.
\newblock In: Proceedings of the thirtieth annual ACM symposium on Theory of
  computing, ACM (1998)  604--613

\bibitem{jegou2011product}
Jegou, H., Douze, M., Schmid, C.:
\newblock Product quantization for nearest neighbor search.
\newblock IEEE transactions on pattern analysis and machine intelligence
  \textbf{33}(1) (2011)  117--128

\bibitem{Holidays}
Jégou, H., Douze, M., Schmid, C.:
\newblock Hamming embedding and weak geometry consistency for large scale image
  search-extended version.
\newblock (2008)

\bibitem{jegouVLAD}
Jégou, H., Douze, M., Schmid, C., Pérez, P.:
\newblock Aggregating local descriptors into a compact image representation.
\newblock CVPR (2010)  3304--3311

\bibitem{kalantidis2014locally}
Kalantidis, Y., Avrithis, Y.:
\newblock Locally optimized product quantization for approximate nearest
  neighbor search.
\newblock In: Proceedings of the IEEE Conference on Computer Vision and Pattern
  Recognition. (2014)  2321--2328

\bibitem{liu2007survey}
Liu, Y., Zhang, D., Lu, G., Ma, W.Y.:
\newblock A survey of content-based image retrieval with high-level semantics.
\newblock Pattern recognition \textbf{40}(1) (2007)  262--282

\bibitem{lv2007multi}
Lv, Q., Josephson, W., Wang, Z., Charikar, M., Li, K.:
\newblock Multi-probe {LSH}: efficient indexing for high-dimensional similarity
  search.
\newblock In: Proceedings of the 33rd international conference on Very large
  data bases, VLDB Endowment (2007)  950--961

\bibitem{maglianiLocVLAD}
Magliani, F., Bidgoli, N.M., Prati, A.:
\newblock A location-aware embedding technique for accurate landmark
  recognition.
\newblock ICDSC (2017)

\bibitem{muja2014scalable}
Muja, M., Lowe, D.G.:
\newblock Scalable nearest neighbor algorithms for high dimensional data.
\newblock IEEE transactions on pattern analysis and machine intelligence
  \textbf{36}(11) (2014)  2227--2240

\bibitem{norouzi2012fast}
Norouzi, M., Punjani, A., Fleet, D.J.:
\newblock Fast search in hamming space with multi-index hashing.
\newblock In: Computer Vision and Pattern Recognition (CVPR), 2012 IEEE
  Conference on, IEEE (2012)  3108--3115

\bibitem{perronnin2010large}
Perronnin, F., Liu, Y., S{\'a}nchez, J., Poirier, H.:
\newblock Large-scale image retrieval with compressed fisher vectors.
\newblock In: Computer Vision and Pattern Recognition (CVPR), 2010 IEEE
  Conference on, IEEE (2010)  3384--3391

\bibitem{Oxford}
Philbin, J., Chum, O., Isard, M., Sivic, J., Zisserman, A.:
\newblock Object retrieval with large vocabularies and fast spatial matching.
\newblock In: Proceedings of the IEEE Conference on Computer Vision and Pattern
  Recognition. (2007)

\bibitem{Paris}
Philbin, J., Chum, O., Isard, M., Sivic, J., Zisserman, A.:
\newblock Lost in quantization: Improving particular object retrieval in alrge
  scale image databases.
\newblock CVPR (2008)

\bibitem{sivicBoW}
Sivic, J., Zisserman, A.:
\newblock Video google: A text retrieval approach to object matching in videos.
\newblock ICCV \textbf{2} (2003)  1470--1477

\bibitem{szegedy2016rethinking}
Szegedy, C., Vanhoucke, V., Ioffe, S., Shlens, J., Wojna, Z.:
\newblock Rethinking the inception architecture for computer vision.
\newblock In: Proceedings of the IEEE Conference on Computer Vision and Pattern
  Recognition. (2016)  2818--2826

\bibitem{tolias2015particular}
Tolias, G., Sicre, R., J{\'e}gou, H.:
\newblock Particular object retrieval with integral max-pooling of {CNN}
  activations.
\newblock arXiv preprint arXiv:1511.05879 (2015)

\bibitem{weiss2009spectral}
Weiss, Y., Torralba, A., Fergus, R.:
\newblock Spectral hashing.
\newblock In: Advances in neural information processing systems. (2009)
  1753--1760

\bibitem{yue2015exploiting}
Yue-Hei~Ng, J., Yang, F., Davis, L.S.:
\newblock Exploiting local features from deep networks for image retrieval.
\newblock In: Proceedings of the IEEE Conference on Computer Vision and Pattern
  Recognition Workshops. (2015)  53--61

\end{thebibliography}
\end{document}